\documentclass[runningheads]{llncs}
\usepackage{graphicx}
\usepackage{amsmath,amssymb} 
\usepackage{color}

\begin{document}
\pagestyle{headings}
\mainmatter

\def\ACCV20SubNumber{680}  

\title{Adversarial Refinement Network for Human Motion Prediction} 
\titlerunning{ARNet for Human Motion Prediction}
%
\author{Xianjin CHAO \inst{1}\orcidID{0000-0002-4020-8223} \and
Yanrui Bin\inst{2}\orcidID{0000-0003-2845-3928} \and
Wenqing Chu\inst{3} \and
Xuan Cao\inst{3} \and
Yanhao Ge\inst{3} \and
Chengjie Wang\inst{3}\and
Jilin Li\inst{3} \and
Feiyue Huang\inst{3} \and
Howard Leung\inst{1}\orcidID{0000-0002-2633-2965}}

\authorrunning{X.Chao et al.}
%

\institute{City University of Hong Kong, Hong Kong, China\\
\email{xjchao2-c@my.cityu.edu.hk}\\
\email{howard@cityu.edu.hk} \and
Key Laboratory of Image Processing and Intelligent Control, School of Artificial Intelligence and Automation, Huazhong University of Science and Technology, Wuhan, China\\
\email{yrbin@hust.edu.cn}\\ \and
Tencent Youtu Lab, Shanghai, China\\
\email{\{wenqingchu, marscao, halege, jasoncjwang, jerolinli, garyhuang\}@tencent.com}}

\maketitle

\begin{abstract}
Human motion prediction aims to predict future 3D skeletal sequences by giving a limited human motion as inputs. 
Two popular methods, recurrent neural networks and feed-forward deep networks, are able to predict rough motion trend, but motion details such as limb movement may be lost.
To predict more accurate future human motion, we propose an Adversarial Refinement Network (ARNet) following a simple yet effective coarse-to-fine mechanism with novel adversarial error augmentation.
Specifically, we take both the historical motion sequences and coarse prediction as input of our cascaded refinement network to predict refined human motion and strengthen the refinement network with adversarial error augmentation. During training, we deliberately introduce the error distribution by learning through the adversarial mechanism among different subjects. In testing, our cascaded refinement network alleviates the prediction error from the coarse predictor resulting in a finer prediction robustly. This adversarial error augmentation provides rich error cases as input to our refinement network, leading to better generalization performance on the testing dataset.
We conduct extensive experiments on three standard benchmark datasets and show that our proposed ARNet outperforms other state-of-the-art methods, especially on challenging aperiodic actions in both short-term and long-term predictions.
\end{abstract}

\section{Introduction}

Given the observed human 3D skeletal sequences, the goal of human motion prediction is to predict plausible and consecutive future human motion which convey abundant clues about the person’s intention, emotion and identity.

Effectively predicting the human motion plays an important role in wide visual computing applications such as human-machine interfaces~\cite{koppula2013anticipating}, smart surveillance~\cite{saquib2018pose}, virtual reality~\cite{elhayek2018fully}, healthcare applications~\cite{yuminaka2016non}, autonomous driving~\cite{paden2016survey} and visual human-object tracking~\cite{gong2011multi}.
However, predicting plausible future human motion is a very challenging task due to the non-linear and highly spatial-temporal dependencies of human body parts during movements~\cite{wang2007gaussian}.Considering the time-series property of human motion sequence, recent deep learning based methods formulated the human motion prediction task as a sequence-to-sequence problem and achieved remarkable progresses by using chain-structured Recurrent Neural Networks (RNNs) to capture the temporal dependencies frame-by-frame among motion sequence.
However, recent literature \cite{tang2018long} indicated that the chain-structured RNNs suffer from error accumulation in temporal modeling and deficiency in spatial dynamic description, leading to problems such as imprecise pose and mean pose in motion prediction.

Feed-forward deep networks ~\cite{mao2019learning} are regarded as alternative solutions for human motion prediction task by learning rich representation from all input motion sequences at once. The holistic reasoning of the human motion sequences leads to more consecutive and plausible predictions than chain-structured RNNs.

Unfortunately, current feed-forward deep networks adopt singe-stage architecture and tend to generate the predicted motion coarsely thus yielding unsatisfactory performance, especially for complex aperiodic actions (e.g., Direction or Greeting in H3.6m dataset). The reason is that it is difficult to guide the network to focus more on detailed information when directly predicting the future human motion from limited input information.  

To address the above issues, we propose a novel Adversarial Refinement Network (ARNet) which resorts to a coarse-to-fine framework. We decompose the human motion prediction problem into two stages: coarse motion prediction and finer motion refinement. By joint reasoning of the input-output space of the coarse predictor, we achieve to take both the historical motion sequences and coarse future prediction as input not just one-sided information to polish the challenging human motion prediction task. The coarse-to-fine design allows the refinement module to concentrate on the complete motion trend brought by the historical input and coarse prediction, which are ignored in previous feed-forward deep networks used for human motion prediction. 

Given different actions performed by diverse persons fed to the refinement network in training and testing, the coarse prediction results tend to be influenced by generalization error, which makes it difficult for the refinement network to obtain the fine prediction robustly. We therefore enhance the refinement network with adversarial error distribution augmentation. During training, we deliberately introduce the error distribution by learning through the adversarial mechanism among different subjects based on the coarse prediction. In testing, our cascaded refinement network alleviates the prediction error from the coarse predictor resulting in a finer prediction. Our adversarial component acts as regularization to 
let our network refine the coarse prediction well.
Different from the previous work \cite{gui2018adversarial} which casts the predictor as a generator and introduces discriminator to validate the prediction results, our adversarial training strategy aims to generate error distribution which acts as implicit regularization for better refinement instead of directly generating the skeleton data as prediction. The error augmentation is achieved by a pair of adversarial learning based generator and discriminator. 

Consequently, the proposed ARNet achieves state-of-the-art results on several standard human motion prediction benchmarks over diverse actions categories, especially over the complicated aperiodic actions as shown in Figure~\ref{fig:human-motion-prediction-task}.

Our contributions are summarized as follows:
\begin{itemize}
	\setlength{\itemsep}{0pt}
	\setlength{\parsep}{0pt}
	\setlength{\parskip}{0.2pt}
	\item We propose a coarse-to-fine framework to decompose the difficult prediction problem into coarse prediction task and  refinement task for more accurate human motion prediction.
	\item We design an adversarial learning strategy to produce reasonable error distribution rather than random noise to optimize the refinement network.
	\item The proposed method is comprehensively evaluated on multiple challenging benchmark datasets and outperforms state-of-the-art methods especially on complicated aperiodic actions.

\end{itemize}


\section{Related Work}
\subsection{Human Motion Prediction}
With the emergence of large scale open human motion capture (mocap) datasets, exploring different deep learning architectures to improve human motion prediction performance on diverse actions has become a new trend. 
Due to the inherent temporal-series nature of motion sequence, the chain-structured Recurrent Neural Networks (RNNs) are natively suitable to process motion sequences. The Encoder-Recurrent-Decoder (ERD) model \cite{fragkiadaki2015recurrent} simultaneously learned the representations and dynamics of human motion. The spatial-temporal graph is later employed in \cite{jain2016structural} to construct the Structural-RNNs (SRNN) model for human motion prediction.
The residual connections in RNN model (RRNN) \cite{martinez2017human} helped the decoder model prior knowledge of input human motion. Tang et al. \cite{tang2018long} adopted the global attention and Modified Highway Unit (MHU) to explore motion contexts for long-term dependencies modeling. 
However, these chain-structured RNNs suffer from either frozen mean pose problems or unnatural motion in predicted sequences because of the weakness of RNNs in both long-term temporal memory and spatial structure description.
Feed-forward deep network as an emerging framework has shown the superiority over chain-structured RNNs. 
Instead of processing input frame by frame like chain-structured RNNs, feed-forward deep networks feed all the frames at once, which is a promising alternative for feature extraction to guarantee the integrity and smoothness of long-term temporal information in human motion prediction \cite{tang2018long}.
In this paper, our ARNet is on the basis of feed-forward deep network.

\subsection{Prediction Refinement}
Refinement approaches learn good feature representation from the coarse results in output space and infer the precise location of joints in a further step by recovering from the previous error, which have achieved promisingly improvement in human pose related work.
Multi-stage refinement network \cite{tome2017lifting} associated the coarse pose estimation and refinement in one go to improve the accuracy of 3D human pose estimation by jointly processing the belief maps of 2D joints and projected 3D joints as the inputs to the next stage. Cascaded Pyramid Network (CPN)  \cite{chen2018cascaded} introduced refinement after the pyramid feature network for sufficient context information mining to handle the occluded and invisible joints estimation problems. Another trend of refinement mechanism performed coarse pose estimation and refinement separately. PoseRefiner\cite{fieraru2018learning} refined the given pose estimation by modelling hard pose cases. Posefix \cite{moon2019posefix} proposed an independent pose refinement network for arbitrary human pose estimator and refined the predicted keypoints based on error statistics prior. Patch-based refinement \cite{wan2019patch} utilised the retain fine details from body part patches to improve the accuracy of 3D pose estimation.  
In contrast to the previous work, we further adopt the benefits of refinement network to deal with the problems in 3D human motion prediction via a creative coarse-to-fine manner.


\subsection{Adversarial Learning}
Inspired by the minimax mechanism of Generative Adversarial Networks (GANs) \cite{goodfellow2014generative}, adversarial learning has been widely adopted to train neural networks~\cite{deng2019irc,balaji2019conditional,vankadari2019unsupervised}. 
Several attempts have been proposed to perform data augmentation in the way of adversarial learning, which mainly rely on the pixel manipulation through image synthesis~\cite{chu2019weakly} or a serious of specific image operations~\cite{zhang2020adversarial}. 
The adversarial learning based data augmentation shows powerful potential for model performance improvement. In \cite{frid2018gan}, the results of image recognition achieved promising improvement due to the image synthesis data augmentation.
In human motion prediction, \cite{gui2018adversarial} adopted a predictor with two discriminators to keep the fidelity and continuity of human motion predicted sequences by adversarial training.
In this work, we introduce an online data augmentation scheme in the motion space to improve generalization and optimize the refinement network.

\begin{figure}[t]
    \centering
      \includegraphics[width=0.95\textwidth]{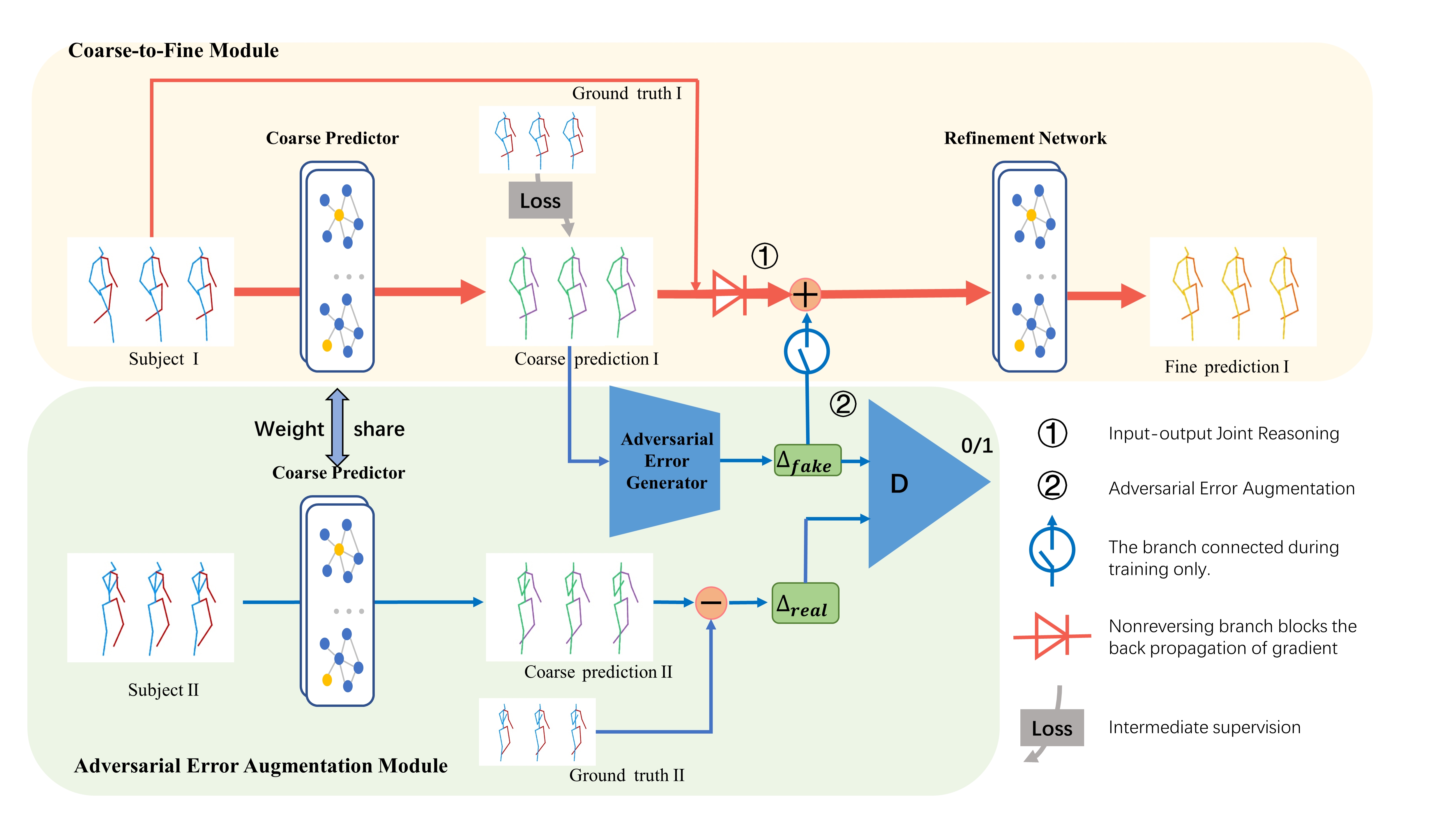}
      \caption{{\bf The overall framework of our ARNet.} The proposed coarse-to-fine module consists of coarse predictor and refinement network as shown in the top part. The bottom part illustrates the dedicated adversarial error augmentation module which consists of coarse predictor with a pair of error generator and discriminator. The observed human motion sequence of Subject I and Subject II are separately fed to the weight-shared coarse predictors to obtain corresponding coarse human motion prediction. Then the generator in the adversarial error augmentation module adopts the coarse prediction of Subject I as the conditional information to generate fake motion error of Subject II in an adversarial manner. After that, the augmented error distribution and the real coarse prediction are both utilised to optimize the refinement network for fine human motion prediction }
      \label{fig:net-structure}
\end{figure}
\section{Methodology}
\subsection{Overall Framework}
The overall framework of our ARNet is shown in Figure~\ref{fig:net-structure}.
The coarse-to-fine module consists of a coarse predictor $\mathcal{P}$ and a refinement network $\mathcal{R}$.
In the context of human motion prediction, given N frames of observed human motion at once, the coarse predictor $\mathcal{P}$ aims to forecast the following T frames of human motion. 
%
%
The input human motion sequences $\mathit{X}=\left \{x_{1},x_{2},...,x_{n}\right\}$ are first fed into the predictor $\mathcal{P}$ to obtain coarse future human motion $\mathit{Y}=\left \{y_{1},y_{2},...,y_{n}\right\}$, where $x_{i}, y_{i}\in\mathbb{R}^{K}$ are $K$ dimensional joint features represented as exponential map of joint angle in each frame.
Then in the adversarial error augmentation module,  we adopt a pair of generator and discriminator to produce fake motion error calculated from the coarse prediction from a person (subject I) and the real motion error from another person (subject II) as the conditional information for the next stage fine prediction.

During training, we deliberately introduce the error distribution by learning through the adversarial mechanism among different subjects based on the coarse prediction. In testing, our cascaded refinement network alleviates the prediction error from the coarse predictor resulting in a finer prediction. 
%

%
%
%
%
%

\subsection{Refinement Network}
Given the input motion sequence, we adopt a Graph Convolutional Network (GCN) \cite{kipf2016semi,yan2018spatial}, a popular feed-forward deep network which is specialized in dealing with the graph structured data, to initially model the spatial-temporal dependencies among the human poses and obtain the coarse human motion prediction.
We construct a $K$ nodes graph $G = (V,E)$, where $V =\left \{v_{i}|i=1,...,K\right \}$ denotes the node set and $E = \left \{e_{i,j}|i,j=1,...,K\right \}$ denotes the edge set.
The main idea of Graph Convolutional Network is that, each $d$ dimensional node representations $H_v^l\in\mathbb{R}^d$ is updated by feature aggregation of all its neighbors defined by the weighted adjacency matrix  $A^l\in\mathbb{R}^{K\times K}$ on the $l$-th Graph Convolutional layer.
Therefore, the spatial structure relationships between the nodes could be fully encoded and the $l$-th Graph Convolution layer outputs a $K \times d$ matrix $H^{l+1} \in\mathbb{R}^d$:
\begin{equation}
H^{l+1} = \sigma(A^lH^lW^l)
\end{equation}
where $\sigma(\cdot)$ denotes an activation function and $W^l \in \mathbb{R}^{d \times \hat{d}}$ denotes the trainable weight matrix. 
The network architecture of our predictor is similar to \cite{mao2019learning}, which is the state-of-the-art feed-forward baseline on human motion prediction.



In order to improve the human motion prediction performance in a further step, we construct a coarse-to-fine framework, which cascades N-stage refinement network on top of the preliminary predictor, to process the complete future information of the output from human motion predictor iteratively.
%
%
Given the input human motion sequences $H_I $, we initially obtain the coarse human motion prediction sequences $H_{\mathcal{P}} = \mathit{f}_p(H_I)$ from the preliminary predictor and forward the fusion of historical and future sequences as the inputs to the refinement network.
As a result, we output the final refined human motion prediction sequences by error correction of initially coarse prediction $H_{\mathcal{R}} = \mathit{f}_r(H_{\mathcal{P}} + H_I)$.
%

%


\subsection{Adversarial Learning Enhanced Refinement Network}
Considering that the human motion sequences collected by different actors in datasets contain variations, especially for complicated aperiodic actions, various error scenarios will occur.
To improve the error-correction ability and robustness of our refinement network, we additionally introduce an adversarial learning mechanism to generate challenging error cases which are fed to the refinement network together with the coarse prediction.
We randomly choose 1 person's actions sequences (Subject II) from the 6 subjects' actions sequences in the training dataset and feed it to the predictor in another branch to get the independent coarse prediction sequences for every epoch as shown in Figure~\ref{fig:net-structure}.
Then the real error is able to be computed from this person's coarse prediction sequences and the corresponding ground-truth.
To augment this person's error cases to the other 5 people, we utilise a generator that produces fake human motion error to fool the discriminator. The discriminator constantly tries to distinguish between real error cases and fake error cases so as to transfer different persons' error to other subjects in the mocap dataset.
This augmentation provides rich error cases as input to our refinement network, leading to better generalization performance on the testing dataset.

We train the networks following the standard GAN pipeline.
During training, the adversarial error generator generates error bias which will be added on the coarse prediction and then fed to refinement network.
The adversarial refinement network effectively learns from the coarse prediction with adversarial error augmentation. During testing, the coarse prediction without added error is fed directly to the adversarial refinement network and get finer prediction as final results.

%

\subsection{Training Loss}
In this section, we describe the training loss functions for different modules. Notably, in order to achieve joint reasoning of the input-output space of the coarse predictor, our ARNet defines the loss function in predictor and refinement network separately to achieve simultaneous supervision. 
Following~\cite{mao2019learning}, we optimize the coarse predictor network parameters with the mean-squared loss, which is denoted as the prediction loss $\mathcal{L_P}$.
Suppose $K$ is the number of joints in each frame, $N$ is the number of input frames and $T$ is the number of predicted frames, then $\mathcal{L_P}$ can be written as:
\begin{equation}
\mathcal{L_P}  = \frac{1}{(N+T)K}\sum_{n=1}^{N+T}\sum_{k=1}^{K}||h^{'}_{k,n}-h_{k,n}||
\end{equation}
where $h_{k,n}$ and $h^{'}_{k,n}$ respectively represent the ground-truth and predicted joint $k$ in frame n.

For the refinement network to produce the refined human motion sequences, we also adopt the mean-squared loss to optimize the network parameters.
The mean-squared loss $\mathcal{L_R}$ can be written as:
\begin{equation}
\mathcal{L_R}  = \frac{1}{(N+T)K}\sum_{n=1}^{N+T}\sum_{k=1}^{K}||h^{''}_{k,n}-h_{k,n}||
\end{equation}
where $h_{k,n}$ indicates the ground-truth joint in frame $n$, $h^{''}_{k,n}$ is the refined corresponding joint. Our refiner is trained by minimizing the loss function.

The goal of our refinement network is to refine the coarse human motion prediction by utilizing the sequence-level refinement with the adversarial learning based error distribution augmentation. We utilise the minimax mechanism of adversarial loss to train the GAN:
\begin{equation}
\mathcal{L_D}  = \boldsymbol{E}[\mathit{log\mathcal{D}(\delta_{real})}]+\boldsymbol{E}[\mathit{log(1-\mathcal{D}(\mathcal{G}(\delta_{fake})}]
\end{equation}
\begin{equation}
\mathcal{L_G}  = \boldsymbol{E}[\mathit{log(1-\mathcal{D}(\mathcal{G}(\delta_{fake})}]
\end{equation}
where $\mathcal{L_D}$ denotes the discriminator loss, $\mathcal{L_G}$ is the generator loss, and $\boldsymbol{\mathbf{\mathit{\delta}}}$ represents the error distribution.

In summary, we gather the predictor and refinement network together to train the whole network in an end-to-end way. As we adopt the adversarial refinement network behind the coarse predictor, the objective function consists of two parts:
\begin{equation}
\mathcal{L}  = \mathcal{L_P} + \boldsymbol{s} * \mathcal{L_R} 
\end{equation}
where $\mathcal{L_P}$ denotes the prediction loss, $\mathcal{L_R}$ denotes the refinement loss, and the number of refinement stage $\boldsymbol{s}$ used in our adversarial refinement network will be shown in the ablation studies.






\section{Experiments}
\subsection{Datasets and Evaluation Metrics}
\subsubsection{H3.6m Dataset.}
Human 3.6 Million (H3.6m) dataset \cite{h36m_pami} is the largest and most challenging mocap dataset which has 15 different daily actions performed by 7 males and females, including not only simple periodic actions such as walking and eating, but also complex aperiodic actions such as discussion and purchase. Following previous methods~\cite{LiZLL18,mao2019learning}, the proposed algorithm is trained on subject 1,6,7,8,9,11 and tested on subject 5. There are 25 frames per second and each frame consists of a skeleton of 32 joints. Except for removing the global translations and rotations, some of the joints that do not move (\textit{i.e.}, joints that do not bend)  will be ignored as previous work \cite{mao2019learning}. 

\subsubsection{CMU-Mocap Dataset.}
To be more convincing, we also conduct experiments on the CMU-Mocap dataset~\cite{LiZLL18}.
In order to achieve fair comparisons, we employ the same experimental settings as \cite{LiZLL18,mao2019learning}, including the pre-processing, data representation and training/testing splits.

\subsubsection{3DPW Dataset.}
Recently, the 3D Pose in the Wild dataset (3DPW)~\cite{vonMarcard2018} is released which contains around 51k frames with 3D annotations. The dataset is challenging as the scenarios are composed of indoor and outdoor activities. We follow \cite{vonMarcard2018,mao2019learning} to split the dataset for comparable experimental results.

\subsubsection{Evaluation Metrics.}
In order to make fair and comprehensive comparisons with previous work, we adopt the Mean Angle Error (MAE) between the predicted frames and the ground-truth frames in the angle space as the quantitative evaluation and visualize the prediction as the qualitative evaluation, which are the common evaluation metrics in human motion prediction~\cite{mao2019learning}.

\subsection{Implementation Details}
The proposed algorithm is implemented on Pytorch \cite{paszke2017automatic} and trained on a NVIDIA Tesla V100 GPU. We adopted the Adam \cite{kingma2014adam} optimizer to train our model for about 50 epochs. The learning rate was set to 0.002 and the batch size was 256. 
To tackle the long-term temporal memory problems, we encode the complete time series by using Discrete Cosine Transform (DCT) \cite{akhter2009nonrigid} and discard the high-frequency jittering to maintain complete expression and smooth consistency of temporal domain information \cite{mao2019learning} at one time.

\subsection{Quantitative Comparisons}
We conduct quantitative comparisons on three human mocap datasets including H3.6m, 3DPW and CMU-Mocap between our ARNet and the state-of-the-art baselines. For fair comparisons with previous work \cite{dong2019retrospecting,gui2018adversarial,LiZLL18,mao2019learning,martinez2017human}, we feed 10 frames as inputs to predict the future 10 frames (400ms) for short-term prediction and the future 25 frames (1000ms) for long-term prediction. 

\begin{table}[t]
    \centering
    \caption{Short-term (80ms,160ms,320ms,400ms) human motion prediction measured in mean angle error (MAE) over 15 actions on H3.6m dataset}
    \resizebox{1.0\textwidth}{!}{
    \begin{tabular}{lcccc|cccc|cccc|cccc}
        & \multicolumn{4}{c}{Walking} & \multicolumn{4}{c}{Eating} & \multicolumn{4}{c}{Smoking} & \multicolumn{4}{c}{Discussion} \\
        \multicolumn{1}{c}{milliseconds} & 80 & 160 & 320 & 400 & 80 & 160 & 320 & 400  & 80 & 160 & 320 & 400  & 80 & 160 & 320 & 400 \\ 
        \hline
        Zero-velocity \cite{martinez2017human} & 0.39 & 0.68 & 0.99 & 1.15  & 0.27 & 0.48 & 0.73 & 0.86 & 0.26 & 0.48 & 0.97 & 0.95  & 0.31 & 0.67 & 0.94 & 1.04 \\
        Residual sup. \cite{martinez2017human} & 0.28 & 0.49 & 0.72 & 0.81  & 0.23 & 0.39 & 0.62 & 0.76  & 0.33 & 0.61 & 1.05 & 1.15 & 0.31 & 0.68 & 1.01 & 1.09 \\
        convSeq2Seq \cite{LiZLL18} & 0.33 & 0.54 & 0.68 & 0.73  & 0.22 & 0.36 & 0.58 & 0.71  & 0.26 & 0.49 & 0.96 & 0.92  & 0.32 & 0.67 & 0.94 & 1.01 \\
        Retrospec \cite{dong2019retrospecting}& 0.28 & 0.45 & 0.62 & 0.68  & 0.21 & 0.34 & 0.53 & 0.68  & 0.26 & 0.50 & 0.96 & 0.93  & 0.29 & 0.64 & 0.90  & 0.96 \\ 
        AGED  \cite{gui2018adversarial} & 0.22 & 0.36 & 0.55 & 0.67  & 0.17 & \textbf{0.28} & 0.51 & 0.64 & 0.27 & 0.43 & \textbf{0.82} & 0.84  & 0.27 & 0.56 & \textbf{0.76} & \textbf{0.83} \\ 
        LTraiJ  \cite{mao2019learning} & \textbf{0.18} & \textbf{0.31} & \textbf{0.49} & 0.56  & \textbf{0.16} & 0.29 & 0.50 & 0.62 & \textbf{0.22} & \textbf{0.41} & 0.86 & \textbf{0.80}  & \textbf{0.20} & \textbf{0.51} & 0.77 & 0.85\\ \hline
        \multicolumn{1}{c}{ARNet (Ours)} & \textbf{0.18} & \textbf{0.31} & \textbf{0.49} & \textbf{0.55} & \textbf{0.16} & \textbf{0.28} & \textbf{0.49} & \textbf{0.61}   & \textbf{0.22} & 0.42& 0.86 & 0.81 &  \textbf{0.20} & \textbf{0.51} & 0.81 & 0.89 \\
        \hline
        \\
        & \multicolumn{4}{c}{Direction} & \multicolumn{4}{c}{Greeting} & \multicolumn{4}{c}{Phoning} & \multicolumn{4}{c}{Posing} \\
        \multicolumn{1}{c}{milliseconds} & 80 & 160 & 320 & 400  & 80 & 160 & 320 & 400 & 80 & 160 & 320 & 400  & 80 & 160 & 320 & 400 \\ 
        \hline
        Zero-velocity \cite{martinez2017human} & 0.39 & 0.59 & 0.79 & 0.89  & 0.54 & 0.89 &1.30 & 1.49 & 0.64 & 1.21 &1.65 &1.83  & 0.28 & 0.57 & 1.13 & 1.37  \\
        Residual sup. \cite{martinez2017human} & 0.26 & 0.47 & 0.72 & 0.84  & 0.75 &1.17 & 1.74 & 1.83 &0.23 &0.43 &0.69 &0.82   & 0.36 & 0.71 & 1.22 & 1.48  \\
        convSeq2Seq \cite{LiZLL18} & 0.39 & 0.60 & 0.80 & 0.91  & 0.51 & 0.82 & 1.21 & 1.38 & 0.59 & 1.13 & 1.51 & 1.65 & 0.29 & 0.60 & 1.12 & 1.37 \\
        Retrospec \cite{dong2019retrospecting}& 0.40 & 0.61 & 0.77 & 0.86 & 0.52 & 0.86 & 1.26 & 1.43  & 0.59 & 1.11 & 1.47 & 1.59  & 0.26 & 0.54 & 1.14 & 1.41 \\ 
        AGED  \cite{gui2018adversarial} & \textbf{0.23}& \textbf{0.39} & \textbf{0.63}& \textbf{0.69}  & 0.56 & 0.81 & 1.30 & 1.46 & \textbf{0.19} &\textbf{0.34} &\textbf{0.50} & \textbf{0.68}  & 0.31& 0.58 & 1.12 & 1.34 \\ 
        LTraiJ \cite{mao2019learning} & 0.26 & 0.45 & 0.71 & 0.79  & 0.36 & 0.60 & 0.95 & 1.13 & 0.53 & 1.02 & 1.35 & 1.48 & 0.19 & 0.44 & 1.01 & 1.24 \\\hline
        \multicolumn{1}{c}{ARNet (Ours)} & \textbf{0.23} & 0.43 &0.65 & 0.75 &  \textbf{0.32} & \textbf{0.55} & \textbf{0.90} & \textbf{1.09}   & 0.51 & 0.99& 1.28 & 1.40 &  \textbf{0.17} & \textbf{0.43} & \textbf{0.97} & \textbf{1.20} \\
        \hline
        \\
        & \multicolumn{4}{c}{Purchases} & \multicolumn{4}{c}{Sitting} & \multicolumn{4}{c}{Sitting Down} & \multicolumn{4}{c}{Taking Photo} \\
        \multicolumn{1}{c}{milliseconds} & 80 & 160 & 320 & 400  & 80 & 160 & 320 & 400  & 80 & 160 & 320 & 400  & 80 & 160 & 320 & 400 \\ \hline
        Zero-velocity \cite{martinez2017human} & 0.62 & 0.88 & 1.19 & 1.27 & 0.40 & 1.63 & 1.02 & 1.18  & 0.39 & 0.74 & 1.07 & 1.19  & 0.25 & 0.51 & 0.79 & 0.92  \\
        Residual sup. \cite{martinez2017human}  & 0.51 & 0.97 & 1.07 & 1.16  & 0.41 & 1.05 & 1.49 & 1.63  & 0.39 & 0.81 & 1.40 & 1.62  & 0.24 & 0.51 & 0.90 & 1.05 \\
        convSeq2Seq \cite{LiZLL18}& 0.63 & 0.91 & 1.19 & 1.29 & 0.39 & 0.61 & 1.02 &1.18  & 0.41 & 0.78 & 1.16 & 1.31  & 0.23 & 0.49 & 0.88 & 1.06 \\
        Retrospec \cite{dong2019retrospecting}& 0.59 & 0.84 & 1.14 & 1.19  & 0.40 & 0.64 & 1.04 & 1.22  & 0.41 & 0.77 & 1.14 & 1.29  & 0.27 & 0.52 & 0.80 & 0.92 \\ 
        AGED  \cite{gui2018adversarial}& 0.46& 0.78& 1.01& \textbf{1.07}  & 0.41 & 0.76 & 1.05 & 1.19  & 0.33 & 0.62 & 0.98 & 1.10  & 0.23 & 0.48 & 0.81 & 0.95 \\ 
        LTraiJ \cite{mao2019learning} & 0.43 & 0.65 & 1.05 & 1.13  & 0.29 & 0.45  & \textbf{0.80} & \textbf{0.97}  & 0.30 & \textbf{0.61} & 0.90 & 1.00  & 0.14 & 0.34 & 0.58 & 0.70 \\\hline
        \multicolumn{1}{c}{ARNet (Ours)} & \textbf{0.36} & \textbf{0.60} & \textbf{1.00} & 1.11  &  \textbf{0.27} & \textbf{0.44} & \textbf{0.80} & \textbf{0.97} &  \textbf{0.29} & \textbf{0.61}& \textbf{0.87} & \textbf{0.97}  &  \textbf{0.13} & \textbf{0.33} & \textbf{0.55} & \textbf{0.67} \\
        \hline
        \\
        & \multicolumn{4}{c}{Waiting} & \multicolumn{4}{c}{Walking Dog} & \multicolumn{4}{c}{Walking Together} & \multicolumn{4}{c}{Average} \\
        \multicolumn{1}{c}{milliseconds} & 80 & 160 & 320 & 400 & 80 & 160 & 320 & 400  & 80 & 160 & 320 & 400  & 80 & 160 & 320 & 400  \\ 
        \hline
        Zero-velocity \cite{martinez2017human} & 0.34 & 0.67 & 1.22 & 1.47  & 0.60 & 0.98 & 1.36 & 1.50  & 0.33 & 0.66 & 0.94 & 0.99  & 0.40 & 0.78 & 1.07 & 1.21  \\
        Residual sup. \cite{martinez2017human}  & 0.28 & 0.53 & 1.02 & 1.14  & 0.56 & 0.91 & 1.26 & 1.40  & 0.31 & 0.58 & 0.87 & 0.91  & 0.36 & 0.67 & 1.02 & 1.15 \\
        convSeq2Seq \cite{LiZLL18}&0.30 & 0.62 & 1.09 & 1.30  & 0.59 & 1.00 & 1.32 & 1.44  & 0.27 & 0.52 & 0.71 & 0.74  & 0.38 & 0.68 & 1.01 & 1.13 \\
        Retrospec \cite{dong2019retrospecting}& 0.33 & 0.65 & 1.12 & 1.30  & 0.53 & 0.87 & 1.16 & 1.33  & 0.28 & 0.52 & 0.68 & 0.71  & 0.37 & 0.66 & 0.98 & 1.10 \\ 
        AGED  \cite{gui2018adversarial}& 0.24 & 0.50 & 1.02 & \textbf{1.13}  & 0.50 & 0.81 & 1.15 & \textbf{1.27}  & 0.23 & 0.41 & 0.56 & 0.62  & 0.31 & 0.54 & 0.85 & 0.97 \\ 
        LTraiJ \cite{mao2019learning} & 0.23 & 0.50 & 0.91 & 1.14  & 0.46 & 0.79 & 1.12 & 1.29  & 0.15 & 0.34 & \textbf{0.52} & \textbf{0.57}  & {0.27} & {0.51} & {0.83} & {0.95}  \\\hline
        \multicolumn{1}{c}{ARNet (Ours)} & \textbf{0.22} & \textbf{0.48} & \textbf{0.90} & \textbf{1.13} &  \textbf{0.45} & \textbf{0.78} & \textbf{1.11}& \textbf{1.27}   & \textbf{0.13} & \textbf{0.33} &{0.53} & {0.58}  &  \textbf{0.25} & \textbf{0.49} & \textbf{0.80} & \textbf{0.92}\\
        \hline
    \end{tabular}
    }
    \label{table-short-motion}
\end{table}


\subsubsection{Short-term Prediction on H3.6m.}
H3.6m is the most challenging dataset for human motion prediction. Table~\ref{table-short-motion} shows the quantitative comparisons for short-term human motion prediction between our ARNet and a series of baselines including Zero-velocity \cite{martinez2017human}, RRNN\cite{martinez2017human}, convSeq2Seq\cite{LiZLL18}, Retrospec\cite{dong2019retrospecting}, AGED \cite{gui2018adversarial} and LTraiJ \cite{mao2019learning} on H3.6m dataset. We computed the mean angle error (MAE) on 15 actions by measuring the euclidean distance between the ground-truth and prediction at 80ms, 160ms, 320ms, 400ms for short-term evaluation. The results in bold show that our method outperforms both of the state-of-the-art chain-structured baseline AGED and the feed-forward baseline LTraiJ. 

Compared with the state-of-the-art feed-forward baseline LTraiJ \cite{mao2019learning}, in Table~\ref{table-short-motion}, the proposed ARNet clearly outperforms the feed-forward baseline LTraiJ on average for short-term human motion prediction. Different from LTraiJ which adopts the single-stage predictor without refinement network, our ARNet obtains better performance especially on aperiodic actions (e.g. Directions, Greeting, Phoning and so on). It is difficult to model this type of actions which involved multiple small movements and high acceleration during human motion especially at the end of human limbs. In addition, due to the stable change of periodic behavior, the traditional feed-forward deep network can also achieve competitive results on periodic actions (such as walking, eating and smoking), but we note that our ARNet further improves the accuracy of prediction. The results validate that the coarse-to-fine design enables our  ARNet to correct the error joints in human motion prediction and outperform the existing feed-forward baseline on almost all actions.

Compared with the state-of-the-art chain-structured baseline AGED  \cite{gui2018adversarial}, which utilises chain-structured RNNs as the predictor with two different discriminators, our ARNet still outperforms it on almost all action categories for short-term human motion prediction within 400ms as shown in Table ~\ref{table-short-motion}. The results show the superiority of our ARNet over the best performing chain-structured methods for short-term human motion prediction tasks.

\begin{figure}[t]
 \centering
   \includegraphics[width=0.75\textwidth]{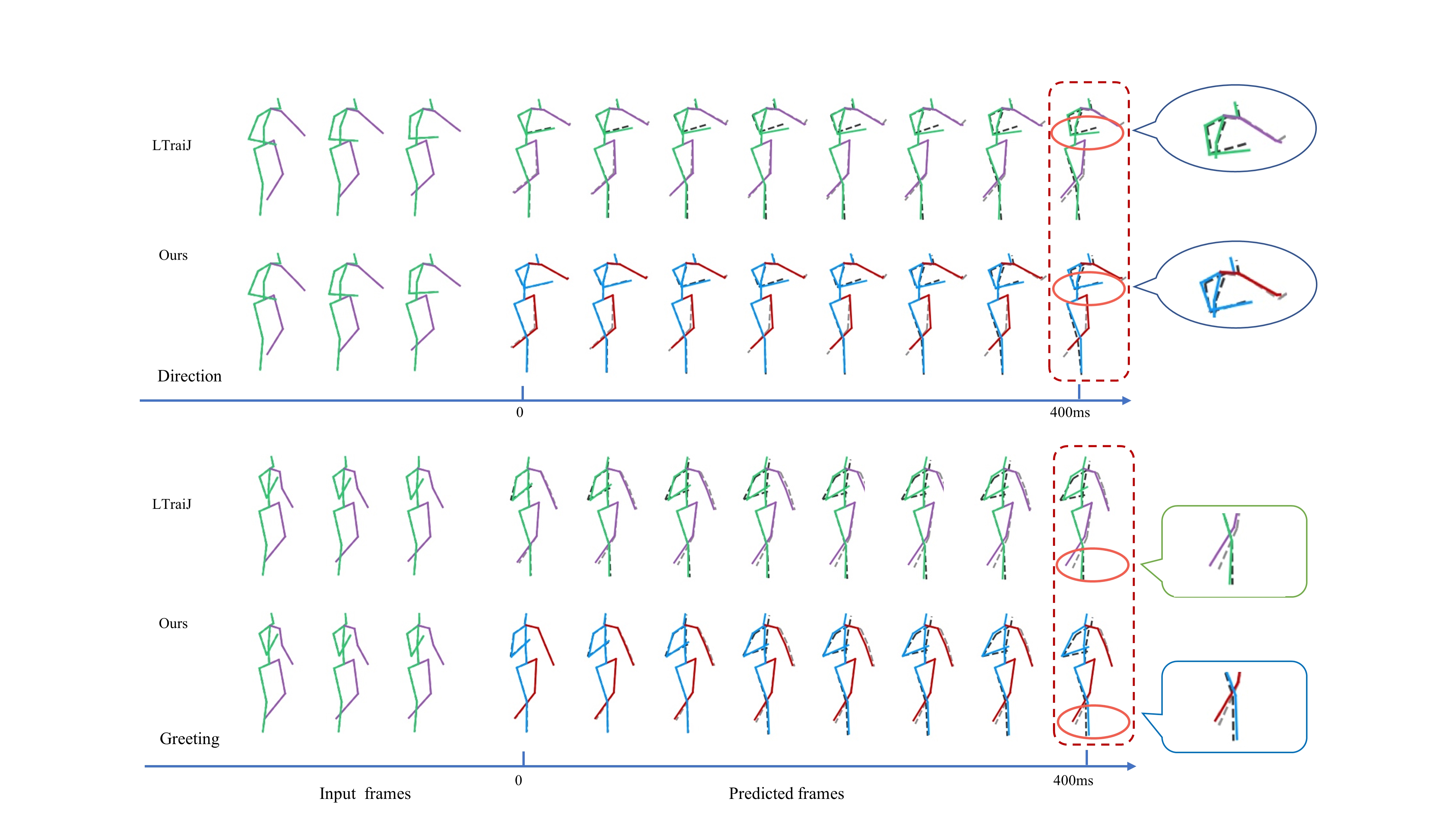}
    \caption{{\bf Visual comparisons for short-term human motion prediction on H3.6m dataset.} We compare our proposed ARNet with the state-of-the-art feed-forward baseline LTraiJ ~\cite{mao2019learning} which is the best performing method for short-term prediction (400ms). The left few frames represent the input human motion sequence. From top to bottom, we show the final predictions obtained by the feed-forward baseline LTraiJ represented as green-purple skeletons and our proposed ARNet represented as red-blue skeletons respectively on two challenging aperiod actions (e.g.,Direction and Greeting). Marked in red circles, our predictions better match the ground-truth shown as the gray dotted skeletons}
    \label{fig:human-motion-prediction-task}
\end{figure}

\begin{table}[h]
    \centering
    \caption{ Long-term (560ms, 1000ms) human motion prediction on H3.6m dataset}
    \resizebox{0.8\textwidth}{!}{%
    \begin{tabular}{lcc|cc|cc|cc|cc}
        ~ & \multicolumn{2}{c}{Walking} & \multicolumn{2}{c}{Eating} & \multicolumn{2}{c}{Smoking} & \multicolumn{2}{c}{Discussion}&\multicolumn{2}{c}{Average} \\
        \multicolumn{1}{c}{milliseconds} & 560 & 1000 & 560 & 1000 & 560 & 1000 & 560 & 1000& 560 & 1000 \\ 
        \hline
        Zero-velocity \cite{martinez2017human} & 1.35 & 1.32 & 1.04 & 1.38 & 1.02 & 1.69 & 1.41 & 1.96 & 1.21 & 1.59\\
        Residual sup.  \cite{martinez2017human} & 0.93 & 1.03 & 0.95 & 1.08 & 1.25 & 1.50 & 1.43 & 1.69 & 1.14 & 1.33\\
        AGED  \cite{gui2018adversarial} & 0.78 & 0.91 & 0.86& \textbf{0.93} & 1.06 & \textbf{1.21} & \textbf{1.25} & \textbf{1.30}& 0.99 & \textbf{1.09}\\
        Retrospec \cite{dong2019retrospecting} & NA & 0.79 & NA & 1.16 & NA & 1.71 & NA & 1.72 & NA & 1.35\\
        LTraiJ \cite{mao2019learning} & \textbf{0.65} & \textbf{0.67} & 0.76 & 1.12 & 0.87 & 1.57 & 1.33 & 1.70 & 0.90 & 1.27\\
        \hline
        ARNet (Ours) & \textbf{0.65} & 0.69 & \textbf{0.72} & 1.07 & \textbf{0.86} & 1.51 &  \textbf{1.25} & 1.68 &\textbf{0.88} &1.24 \\
        \hline
    \end{tabular}
    }
    \label{table-long-motion}
\end{table}

\subsubsection{Long-term Prediction on H3.6m.}
Additionlly, we also quantitatively evaluate the long-term prediction performance of our proposed ARNet at 560ms and 1000ms as shown in Table~\ref{table-long-motion}. The results measured in MAE demonstrate that our method still outperforms the state-of-art feed-forward baseline LTraiJ \cite{mao2019learning} in long-term human motion prediction on almost action categories as shown in bold. Nevertheless, the MAE of the chain-structured AGED \cite{gui2018adversarial} is lower than ours in 1000 milliseconds. We will further examine the results by visualizing the motion sequences obtained by our proposed ARNet and the chain-structured baseline AGED in the later section to provide a qualitative comparison.

\subsubsection{3DPW $\&$ CMU-Mocap.}
We also conduct experiments on other two human mocap datasets to prove the robustness of our method. Table~\ref{table-short-long-3dpw} shows that our method consistently achieves promising improvements compared with other baselines on 3DPW dataset which contains indoor and outdoor activities for both short-term and long-term human motion predictions. As for CMU-Mocap dataset, the results in Table~\ref{table-short-motion-cmu} illustrate that our method has better performance on almost action types and outperforms the state-of-the-art methods on average. 

\begin{table}[h]
    \centering
    \caption{ Short-term and long-term human motion predictions on 3DPW dataset}
    \resizebox{0.5\textwidth}{!}{%
    \begin{tabular}{l|ccccc}
        \multicolumn{1}{c}{milliseconds} & 200 & 400 & 600 & 800 & 1000 \\ 
        \hline
        Residual sup.~\cite{martinez2017human} & 1.85 & 2.37 & 2.46 & 2.51 & 2.53 \\
        convSeq2Seq~\cite{LiZLL18} & 1.24 & 1.85 &  2.13 & 2.23 & 2.26 \\
        LTraiJ \cite{mao2019learning} & 0.64 & \textbf{0.95} & 1.12 & 1.22 & 1.27 \\
        \hline
        ARNet (Ours) & \textbf{0.62} & \textbf{0.95} & \textbf{1.11} & \textbf{1.20} & \textbf{1.25} \\
        \hline
    \end{tabular}
    }
    \label{table-short-long-3dpw}
\end{table}

\begin{table}[h]
    \centering
    \caption{Short-term and long-term human motion predictions on CMU-Mocap dataset}
    \resizebox{\textwidth}{!}{%
    \begin{tabular}{lccccc|ccccc|ccccc}
        & \multicolumn{5}{c}{Basketball} & \multicolumn{5}{c}{Basketball Signal} & \multicolumn{5}{c}{Directing Traffic}  \\
        milliseconds& 80 & 160 & 320 & 400 & 1000 & 80 & 160 & 320 & 400 & 1000 & 80 & 160 & 320 & 400 & 1000 \\ 
        \hline
        Residual sup. \cite{martinez2017human} & 0.50 & 0.80 & 1.27 & 1.45 & 1.78 & 0.41 & 0.76 & 1.32 & 1.54 & 2.15 & 0.33 & 0.59 & 0.93 & 1.10 & 2.05  \\
        convSeq2Seq \cite{LiZLL18} & 0.37 & 0.62 & 1.07 & 1.18 & 1.95 & 0.32 & 0.59 & 1.04 & 1.24 & 1.96 & 0.25 & 0.56 & 0.89 & 1.00 & 2.04  \\
        LTraiJ \cite{mao2019learning} & 0.33 & 0.52 & 0.89 & \textbf{1.06} & \textbf{1.71} & 0.11 & 0.20 & 0.41 & 0.53 & 1.00 & 0.15 & 0.32 & 0.52 & 0.60 & 2.00  \\
        \hline
        \multicolumn{1}{l}{ARNet (Ours)} & \textbf{0.31} & \textbf{0.48} & \textbf{0.87} & 1.08 & \textbf{1.71} & \textbf{0.10} & \textbf{0.17} & \textbf{0.35} & \textbf{0.48} & \textbf{1.06} & \textbf{0.13} & \textbf{0.28} & \textbf{0.47} & \textbf{0.58}& \textbf{1.80} \\
        \hline
        \\
        & \multicolumn{5}{c}{Jumping} & \multicolumn{5}{c}{Running} & \multicolumn{5}{c}{Soccer}  \\
        milliseconds& 80 & 160 & 320 & 400 & 1000 & 80 & 160 & 320 & 400 & 1000 & 80 & 160 & 320 & 400 & 1000\\ 
        \hline
        Residual sup. \cite{martinez2017human}& 0.33 & 0.50 & 0.66 & 0.75 & 1.00 & 0.29 & 0.51 & 0.88 & 0.99 & 1.72 & 0.56 & 0.88 & 1.77 & 2.02 & 2.4 \\
        convSeq2Seq \cite{LiZLL18}  & \textbf{0.28} & \textbf{0.41} & \textbf{0.52} & \textbf{0.57} & \textbf{0.67} & 0.26 & 0.44 & 0.75 & 0.87 & 1.56 & 0.39 & 0.6 & 1.36 & 1.56 & 2.01 \\
        LTraiJ \cite{mao2019learning} & 0.33 & 0.55 & 0.73 & 0.74 & 0.95 & 0.18 & 0.29 & 0.61 & 0.71 & 1.40 & 0.31 & 0.49 & 1.23 & 1.39 & 1.80 \\
        \hline
        \multicolumn{1}{l}{ARNet (Ours)} & 0.30 & 0.50 & 0.60 & 0.61 & 0.72 & \textbf{0.16} & \textbf{0.26} & \textbf{0.57} & \textbf{0.67} & \textbf{1.22} & \textbf{0.29} & \textbf{0.47} & \textbf{1.21} & \textbf{1.38} & \textbf{1.70} \\
        \hline
        \\
        & \multicolumn{5}{c}{Walking} & \multicolumn{5}{c}{Washwindow} & \multicolumn{5}{c}{Average} \\
        milliseconds & 80 & 160 & 320 & 400 & 1000 & 80 & 160 & 320 & 400 & 1000 & 80 & 160 & 320 & 400 & 1000  \\
        \hline
        Residual sup. \cite{martinez2017human}  & 0.35 & 0.47 & 0.60 & 0.65 & 0.88 & 0.30 & 0.46 & 0.72 & 0.91 & 1.36 & 0.38 & 0.62 & 1.02 & 1.18 & 1.67\\
        convSeq2Seq \cite{LiZLL18} & 0.35 & 0.44 & 0.45 & 0.50 & 0.78 & 0.30 & 0.47 & 0.80 & 1.01 & 1.39 & 0.32 & 0.52 & 0.86 & 0.99 & 1.55\\
        LTraiJ \cite{mao2019learning} & 0.33 & 0.45 & 0.49 & 0.53 & 0.61 &0.22 & 0.33 & 0.57 & 0.75 & 1.20 & 0.25 & 0.39 & 0.68 & 0.79 & 1.33\\
        \hline
        \multicolumn{1}{l}{ARNet (Ours)}  & \textbf{0.32} & \textbf{0.41} & \textbf{0.39} & \textbf{0.41} & \textbf{0.56} & \textbf{0.20} & \textbf{0.27} & \textbf{0.51} & \textbf{0.69} & \textbf{1.07} & \textbf{0.23} & \textbf{0.37} & \textbf{0.65} & \textbf{0.77} & \textbf{1.29} \\
        \hline
    \end{tabular}
    }
    \label{table-short-motion-cmu}
\end{table}

\subsection{Qualitative Visualizations}
\subsubsection{Short-term Prediction on H3.6m.} To evaluate our method qualitatively, we firstly visualize the representative comparisons on Directions and Greeting which belong to challenging aperiodic actions in H3.6m dataset as shown in Figure~\ref{fig:human-motion-prediction-task}. 
Given 10 observed frames for each action as motion seeds, which are represented as green-purple skeletons at the left part, we compare our ARNet represented as red-blue skeletons with the best quantitatively  performing feed-forward baseline LTraiJ \cite{mao2019learning} shown as green-purple skeletons for short-term prediction (400 million seconds) as illustrated in Table~\ref{table-short-motion}. 
The dotted rectangles mark that our predictions better match the ground-truth which is represented as gray dotted skeletons. The qualitative comparison further demonstrates that our ARNet possesses the ideal error-correction ability to generate high-quality prediction, especially for the joints at the end of body which contain multiple small movements on aperiodic actions.

\subsubsection{Long-term Prediction on H3.6m.} 
\begin{figure}[t]
 \centering
  \includegraphics[width=1.0\columnwidth]{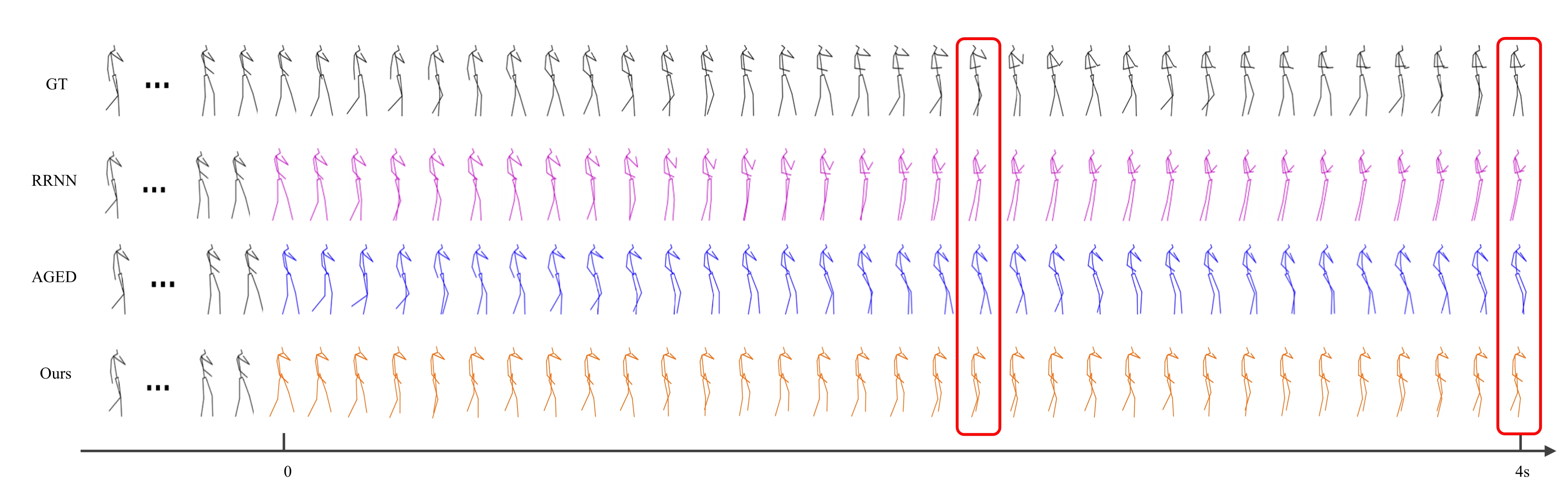}
    \caption{{\bf Visual comparisons for long-term human motion prediction on H3.6m dataset.} From top to bottom,we show the corresponding ground-truth shown in grey skeletons,  the final predictions obtained by RRNN \cite{martinez2017human} , AGED \cite{gui2018adversarial} and our approach on Phoning which belongs to the aperiodic action. The left gray skeletons represent the input motion sequences. Marked in red rectangles, the baseline RRNN converges to mean pose and the baseline AGED drifts away on the foot joints compared with the ground-truth. Our ARNet generates more accurate long-term human motion prediction relatively. Best viewed in color with zoom
    }
    \label{fig:long-term-comparisons}
\end{figure}

Figure~\ref{fig:long-term-comparisons} visualizes the comparisons between chain-structured baselines RRNN \cite{martinez2017human}  and AGED  \cite{martinez2017human} on Phoning, which belongs to aperiodic actions in H3.6m dataset for long-term prediction (4 seconds). As marked by the red rectangles, our proposed ARNet is still able to predict the motion dynamics when the RRNN converges to mean pose. Meanwhile, the AGED drifts away on the foot joints compared with the ground-truth. The visualised results demonstrate that our ARNet outperforms the chain-structured baselines in long-term prediction.

\begin{table}[h]
    \centering
    \caption{Ablation study for refined model design and adversarial training strategy. We compared the results measured in MAE of our model with the 1-stage CoarseNet, the 2-stage CoarseNet without future information as refinement and the 2-stage RefineNet with traditional training strategy on H3.6m dataset }
    \resizebox{\textwidth}{!}{%
    \begin{tabular}{lcccccc|cccccc|cccccc}
        & \multicolumn{6}{c}{Direction} & \multicolumn{6}{c}{Posing} & \multicolumn{6}{c}{Greeting} \\
        milliseconds & 80 & 160 & 320 & 400 & 560 & 1000 & 80 & 160 & 320 & 400 & 560 & 1000 & 80 & 160 & 320 & 400 & 560 & 1000\\ 
        \hline
        1-stage CoarseNet  & 0.26 & 0.45 & 0.71 & 0.79 &0.88 &1.29 &0.19 &0.44 &1.01 &1.24 &1.44 &1.64 &0.36 &0.60 &0.95 &1.13 &1.51 &1.70 \\
        2-stage CoarseNet  & 0.25 & 0.45 & 0.67 & 0.78 &0.88 &1.30 &0.19 &0.46 & 1.01 & 1.26 &1.42 &1.68 & 0.34 & 0.60 &0.94 &1.11 &1.66 &1.92 \\
        2-stage RefineNet  & 0.25 & 0.44 & 0.67 & 0.77 &0.86 &1.27 &0.19 &0.43 & 0.99 & 1.23 &1.42 &1.63 & 0.34 & 0.58 &0.92 &1.10 &1.49 &1.63 \\
        ARNet & \textbf{0.23} & \textbf{0.43} & \textbf{0.65} & \textbf{0.75} & \textbf{0.85} &\textbf{1.23} &\textbf{0.17} &\textbf{0.43} & \textbf{0.97} & \textbf{1.20} &\textbf{1.41} &\textbf{1.60} &\textbf{ 0.31} & \textbf{0.55} &\textbf{0.90} &\textbf{1.08} &\textbf{1.46} &\textbf{1.56} \\ 
        \hline
        \\
        & \multicolumn{6}{c}{Greeting} & \multicolumn{6}{c}{Phoning} & \multicolumn{6}{c}{Average(on 15 actions)}\\
        milliseconds & 80 & 160 & 320 & 400 & 560 & 1000 & 80 & 160 & 320 & 400 & 560 & 1000 & 80 & 160 & 320 & 400 & 560 & 1000  \\ \hline
        1-stage CoarseNet &0.36 &0.60 &0.95 &1.13 &1.51 &1.70 &0.53 &1.02 &1.35 &1.48 &1.45 &1.68 &0.27 &0.51 &0.83 &0.95 &1.18&1.59  \\
        2-stage CoarseNet  & 0.34 & 0.60 &0.94 &1.11 &1.66 &1.92 &0.53 &1.02 &1.34 &1.48 &1.58 &1.98 &0.27&0.52 &0.83 &0.95 &1.20 &1.61  \\
        2-stage RefineNet  & 0.34 & 0.58 & 0.94 & 1.10 &1.48 &1.64 &0.52 &1.01 & 1.33 & 1.46 &1.42 &1.65 & 0.27 & 0.50 &0.82 &0.94 &1.17 &1.58 \\
        ARNet  &\textbf{ 0.31} & \textbf{0.55} &\textbf{0.90} &\textbf{1.08} &\textbf{1.46} &\textbf{1.56} &\textbf{0.50} &\textbf{0.99} &\textbf{1.28} &\textbf{1.40} &\textbf{1.41} &\textbf{1.60} &\textbf{0.25} &\textbf{0.49} &\textbf{0.80} &\textbf{0.92} &\textbf{1.16} &\textbf{1.57}
        \\\hline 
    \end{tabular}
    } 
    \label{refiner_contribution}
\end{table}

\section{Ablation Studies}
\subsection{Different Components in Our ARNet}
In order to verify the effectiveness of the different components in our model, we perform comprehensive ablation studies as shown in Table~\ref{refiner_contribution}. Specifically, we compare our ARNet with three baselines: the 1-stage CoarseNet, the 2-stage CoarseNet without future information as refinement and the 2-stage RefineNet with future information and traditional training strategy. The 1-stage CoarseNet denotes that there only exists single coarse predictor module without other components in the whole framework. We utilize the LTraiJ network \cite{mao2019learning} as our coarse predictor. Due to the coarse-to-fine 2-stage structure of our ARNet, the inference time of our ARNet is 56.2ms, which is slightly longer than the 45.4ms of 1-stage CoarseNets on GPU V100. Moreover, another baseline is the 2-stage CoarseNet without future information refinement, which increase the number of layers by simply cascading two 1-stage CoarseNets, utilise the same training strategy as the single coarse predictor by back-propagating the gradient all the way to the beginning. Although the parameters of our ARNet is same as the 2-stage CoarseNet which is twice that of 1-stage CoarseNets, the results show that stacking multi-layers with traditional training strategy fails to improve the performance in a further step and even achieved worse prediction due to over-fitting occurred in stacked feed-forward deep network. Then, the 2-stage RefineNet without adversarial error augmentation leads to improvement over the previous two baselines. Our adversarial refinement network shows the superior performance compared with single-stage model, 2-stage model without refinement and refinement network without adversarial training strategy.

\subsection{Multi-stage Analysis}
We also evaluate the impact of number of stages adopted in our adversarial refinement model by calculating the MAE over 15 actions. The foregoing results in the Table~\ref{number_stage} indicate that the 2-stage refined model design, in general, utilising the output space of previous stage, is simple enough to learn the rich representation and achieves superior results in most cases. The reason is that concatenating more than 2 stages refinement module faces up over-fitting problems and fails to further improve the human motion prediction performance. Taking the efficiency and simplicity into account, we employ the 2-stage adversarial refinement network as the final model design.


\begin{table}[h]
    \centering
    \caption{Ablation study of adversarial refinement network with different number of stages. We compared the results measured in MAE on H3.6m dataset}
    \resizebox{\textwidth}{!}{%
    \begin{tabular}{ccccccc|cccccc|cccccc}
        & \multicolumn{6}{c}{Direction} & \multicolumn{6}{c}{Posing} & \multicolumn{6}{c}{Greeting} \\
        milliseconds& 80 & 160 & 320 & 400 & 560 & 1000 & 80 & 160 & 320 & 400 & 560 & 1000 & 80 & 160 & 320 & 400 & 560 & 1000 \\ \hline
        2-stage & \textbf{0.23} & \textbf{0.43} & \textbf{0.65} & \textbf{0.75} & 0.85 &\textbf{1.23} &\textbf{0.17} &\textbf{0.43} & \textbf{0.97} & \textbf{1.20} &\textbf{1.41} &\textbf{1.60} & \textbf{0.31} & 0.55 &0.90 & \textbf{1.08} &\textbf{1.46} &\textbf{1.56} \\
        3-stage & 0.25 & 0.46 & 0.64 & 0.75 & \textbf{0.84} &1.50 &0.18 &0.44 & 1.00 & 1.25 &1.71 &2.64 & 0.32 & \textbf{0.54} &\textbf{0.89} &1.12 &1.52 &1.75 \\
        4-stage & 0.25 & 0.46 & 0.68 & 0.77 &1.02 &1.70 &0.19 &0.46 & 1.05& 1.28 &1.86 &3.03 & 0.33 & 0.56 &0.93 &1.15 &1.56 &1.82 \\ 
        \hline 
        \\
        & \multicolumn{6}{c}{Greeting} & \multicolumn{6}{c}{Phoning} & \multicolumn{6}{c}{Average(on 15 actions)}\\
        milliseconds& 80 & 160 & 320 & 400 & 560 & 1000 & 80 & 160 & 320 & 400 & 560 & 1000 & 80 & 160 & 320 & 400 & 560 & 1000 \\ \hline
        2-stage & \textbf{0.31} & 0.55 &0.90 & \textbf{1.08} &\textbf{1.46} &\textbf{1.56} &\textbf{0.50} &\textbf{0.99} &\textbf{1.28} &\textbf{1.40} &\textbf{1.41} &\textbf{1.60} & \textbf{0.25} & \textbf{0.49} & \textbf{0.80} & \textbf{0.92} &\textbf{1.16} &\textbf{1.57} \\
        3-stage & 0.32 & \textbf{0.54} &\textbf{0.89}&1.12 &1.52 &1.75 &0.52 &1.02 &1.36 &1.45 &1.49 &1.80 &0.25 &0.49 &0.83 &0.95 &1.17 &1.58  \\
        4-stage & 0.33 & 0.56 &0.93 &1.15 &1.56 &1.82 &0.52 &0.99 & 1.33 & 1.48 & 1.49 &1.76 &0.27 &0.50 &0.83 &0.95 &1.17 &1.58
        \\ \hline 
    \end{tabular}
    }
    \label{number_stage}
\end{table}

\section{Conclusions}
In this paper, we introduce an Adversarial Refinement Network (ARNet) to forecast more accurate human motion sequence in a coarse-to-fine manner. We adopt a refinement network behind the single-stage coarse predictor to generate finer human motion.
Meanwhile, we utilise an adversarial learning strategy to enhance the generalization ability of the refinement network.
Experimental results on the challenging benchmark H3.6m, CMU-Mocap and 3DPW datasets show that our proposed ARNet outperforms the state-of-the-art approaches in both short-term and long-term predictions especially on the complex aperiodic actions. Our adversarial refinement network shows promising potential for feed-forward deep network to deal with rich representation in a further step on other areas.

~\\
\noindent\textbf{Acknowledgement.}
The work described in this paper was supported by grants from City University of Hong Kong(Project No. 9220077 and 9678139).



\bibliographystyle{splncs}
\bibliography{egbib}

\begin{thebibliography}{10}

\bibitem{koppula2013anticipating}
Koppula, H.S., Saxena, A.:
\newblock Anticipating human activities for reactive robotic response.
\newblock In: IROS. (2013)

\bibitem{saquib2018pose}
Saquib~Sarfraz, M., Schumann, A., Eberle, A., Stiefelhagen, R.:
\newblock A pose-sensitive embedding for person re-identification with expanded
  cross neighborhood re-ranking.
\newblock In: CVPR. (2018)

\bibitem{elhayek2018fully}
Elhayek, A., Kovalenko, O., Murthy, P., Malik, J., Stricker, D.:
\newblock Fully automatic multi-person human motion capture for vr
  applications.
\newblock In: International Conference on Virtual Reality and Augmented
  Reality. (2018)

\bibitem{yuminaka2016non}
Yuminaka, Y., Mori, T., Watanabe, K., Hasegawa, M., Shirakura, K.:
\newblock Non-contact vital sensing systems using a motion capture device:
  medical and healthcare applications.
\newblock In: Key engineering materials. (2016)

\bibitem{paden2016survey}
Paden, B., {\v{C}}{\'a}p, M., Yong, S.Z., Yershov, D., Frazzoli, E.:
\newblock A survey of motion planning and control techniques for self-driving
  urban vehicles.
\newblock IEEE Transactions on intelligent vehicles (2016)

\bibitem{gong2011multi}
Gong, H., Sim, J., Likhachev, M., Shi, J.:
\newblock Multi-hypothesis motion planning for visual object tracking.
\newblock In: ICCV. (2011)

\bibitem{wang2007gaussian}
Wang, J.M., Fleet, D.J., Hertzmann, A.:
\newblock Gaussian process dynamical models for human motion.
\newblock IEEE transactions on pattern analysis and machine intelligence
  \textbf{30} (2007)  283--298

\bibitem{tang2018long}
Tang, Y., Ma, L., Liu, W., Zheng, W.:
\newblock Long-term human motion prediction by modeling motion context and
  enhancing motion dynamic.
\newblock In: IJCAI. (2018)

\bibitem{mao2019learning}
Mao, W., Liu, M., Salzmann, M., Li, H.:
\newblock Learning trajectory dependencies for human motion prediction.
\newblock In: ICCV. (2019)

\bibitem{gui2018adversarial}
Gui, L.Y., Wang, Y.X., Liang, X., Moura, J.M.:
\newblock Adversarial geometry-aware human motion prediction.
\newblock In: ECCV. (2018)

\bibitem{fragkiadaki2015recurrent}
Fragkiadaki, K., Levine, S., Felsen, P., Malik, J.:
\newblock Recurrent network models for human dynamics.
\newblock In: ICCV. (2015)

\bibitem{jain2016structural}
Jain, A., Zamir, A.R., Savarese, S., Saxena, A.:
\newblock Structural-rnn: Deep learning on spatio-temporal graphs.
\newblock In: CVPR. (2016)

\bibitem{martinez2017human}
Martinez, J., Black, M.J., Romero, J.:
\newblock On human motion prediction using recurrent neural networks.
\newblock In: CVPR. (2017)

\bibitem{tome2017lifting}
Tome, D., Russell, C., Agapito, L.:
\newblock Lifting from the deep: Convolutional 3d pose estimation from a single
  image.
\newblock In: Proceedings of the IEEE Conference on Computer Vision and Pattern
  Recognition. (2017)  2500--2509

\bibitem{chen2018cascaded}
Chen, Y., Wang, Z., Peng, Y., Zhang, Z., Yu, G., Sun, J.:
\newblock Cascaded pyramid network for multi-person pose estimation.
\newblock In: Proceedings of the IEEE conference on computer vision and pattern
  recognition. (2018)  7103--7112

\bibitem{fieraru2018learning}
Fieraru, M., Khoreva, A., Pishchulin, L., Schiele, B.:
\newblock Learning to refine human pose estimation.
\newblock In: CVPR-W. (2018)

\bibitem{moon2019posefix}
Moon, G., Chang, J.Y., Lee, K.M.:
\newblock Posefix: Model-agnostic general human pose refinement network.
\newblock In: CVPR. (2019)

\bibitem{wan2019patch}
Wan, Q., Qiu, W., Yuille, A.L.:
\newblock Patch-based 3d human pose refinement.
\newblock arXiv preprint arXiv:1905.08231 (2019)

\bibitem{goodfellow2014generative}
Goodfellow, I., Pouget-Abadie, J., Mirza, M., Xu, B., Warde-Farley, D., Ozair,
  S., Courville, A., Bengio, Y.:
\newblock Generative adversarial nets.
\newblock In: NIPS. (2014)

\bibitem{deng2019irc}
Deng, K., Fei, T., Huang, X., Peng, Y.:
\newblock Irc-gan: introspective recurrent convolutional gan for text-to-video
  generation.
\newblock In: IJCAI. (2019)

\bibitem{balaji2019conditional}
Balaji, Y., Min, M.R., Bai, B., Chellappa, R., Graf, H.P.:
\newblock Conditional gan with discriminative filter generation for
  text-to-video synthesis.
\newblock In: IJCAI. (2019)

\bibitem{vankadari2019unsupervised}
Vankadari, M., Kumar, S., Majumder, A., Das, K.:
\newblock Unsupervised learning of monocular depth and ego-motion using
  conditional patchgans.
\newblock In: IJCAI. (2019)

\bibitem{chu2019weakly}
Chu, W., Hung, W.C., Tsai, Y.H., Cai, D., Yang, M.H.:
\newblock Weakly-supervised caricature face parsing through domain adaptation.
\newblock In: ICIP. (2019)

\bibitem{zhang2020adversarial}
Zhang, X., Wang, Q., Zhang, J., Zhong, Z.:
\newblock Adversarial autoaugment.
\newblock In: ICLR. (2020)

\bibitem{frid2018gan}
Frid-Adar, M., Klang, E., Amitai, M., Goldberger, J., Greenspan, H.:
\newblock Gan-based data augmentation for improved liver lesion classification.
\newblock (2018)

\bibitem{kipf2016semi}
Kipf, T.N., Welling, M.:
\newblock Semi-supervised classification with graph convolutional networks.
\newblock (2017)

\bibitem{yan2018spatial}
Yan, S., Xiong, Y., Lin, D.:
\newblock Spatial temporal graph convolutional networks for skeleton-based
  action recognition.
\newblock In: Thirty-second AAAI conference on artificial intelligence. (2018)

\bibitem{h36m_pami}
Ionescu, C., Papava, D., Olaru, V., Sminchisescu, C.:
\newblock Human3.6m: Large scale datasets and predictive methods for 3d human
  sensing in natural environments.
\newblock TPAMI \textbf{36} (2014)  1325--1339

\bibitem{LiZLL18}
Li, C., Zhang, Z., Lee, W.S., Lee, G.H.:
\newblock Convolutional sequence to sequence model for human dynamics.
\newblock In: CVPR. (2018)

\bibitem{vonMarcard2018}
von Marcard, T., Henschel, R., Black, M., Rosenhahn, B., Pons-Moll, G.:
\newblock Recovering accurate 3d human pose in the wild using imus and a moving
  camera.
\newblock In: ECCV. (2018)

\bibitem{paszke2017automatic}
Paszke, A., Gross, S., Chintala, S., Chanan, G., Yang, E., DeVito, Z., Lin, Z.,
  Desmaison, A., Antiga, L., Lerer, A.:
\newblock Automatic differentiation in pytorch.
\newblock In: NIPS-W. (2017)

\bibitem{kingma2014adam}
Kingma, D.P., Ba, J.:
\newblock Adam: A method for stochastic optimization.
\newblock arXiv preprint arXiv:1412.6980 (2014)

\bibitem{akhter2009nonrigid}
Akhter, I., Sheikh, Y., Khan, S., Kanade, T.:
\newblock Nonrigid structure from motion in trajectory space.
\newblock In: NIPS. (2009)

\bibitem{dong2019retrospecting}
Dong, M., Xu, C.:
\newblock On retrospecting human dynamics with attention.
\newblock In: IJCAI. (2019)

\end{thebibliography}

\end{document}